\documentclass{article}

\usepackage{spconf,amsmath,graphicx}
\usepackage{amsfonts}
\usepackage{amsthm}
\usepackage{array}
\usepackage[caption=false, font=small, labelfont=rm, textfont=rm]{subfig}
\usepackage{textcomp}
\usepackage{url}
\usepackage{cite}
\usepackage[colorlinks, linkcolor=black, urlcolor=black, citecolor=black]{hyperref}
\usepackage{color}
\usepackage{makecell}
\usepackage{booktabs}
\usepackage{balance}
\usepackage{soul, xcolor}
\usepackage{multirow}
\usepackage[normalem]{ulem}
\useunder{\uline}{\ul}{}

\allowdisplaybreaks[4]

\newtheorem{assumption}{\bf Assumption}

\newtheorem{problem}{\bf Problem}
\newtheorem{proposition}{\bf Proposition}
\newtheorem{definition}{\bf Definition}

\title{Variational Connectionist Temporal Classification for Order-Preserving Sequence Modeling}
\name{
	Zheng Nan$^{\star}$ \qquad
	Ting Dang$^{\star \dagger \ddagger}$ \qquad
	Vidhyasaharan Sethu$^{\star}$ \qquad
	Beena Ahmed$^{\star}$
    \thanks{
        This research has been supported by an Australian Government Research Training Program (RTP) Scholarship.
    }
}
\address{
	$^{\star}$School of Electrical Engineering and Telecommunications, University of New South Wales, Australia\\
	$^{\dagger}$Nokia Bell Labs, UK\\
	$^{\ddagger}$Department of Computer Science and Technology, University of Cambridge, Cambridge, UK\\
}

\begin{document}
\ninept
\maketitle
\begin{abstract}
Connectionist temporal classification (CTC) is commonly adopted for sequence modeling tasks like speech recognition, where it is necessary to preserve order between the input and target sequences. However, CTC is only applied to deterministic sequence models, where the latent space is discontinuous and sparse, which in turn makes them less capable of handling data variability when compared to variational models. In this paper, we integrate CTC with a variational model and derive loss functions that can be used to train more generalizable sequence models that preserve order. Specifically, we derive two versions of the novel variational CTC based on two reasonable assumptions, the first being that the variational latent variables at each time step are conditionally independent; and the second being that these latent variables are Markovian. We show that both loss functions allow direct optimization of the variational lower bound for the model log-likelihood, and present computationally tractable forms for implementing them.

\end{abstract}

\section{Introduction}\label{sec.introduction}

\noindent Many real-world applications such as speech recognition \cite{graves2013speech, liu2023enhancing}, speech synthesis \cite{kim2021conditional}, handwriting recognition \cite{chao2020variational}, 
etc., involve sequence modeling that estimates the mapping function between two sequences. When the input sequence and the target sequence have the same length (i.e., {they are \it aligned}), recurrent neural networks (RNNs) generally show great promise in modeling the mapping function \cite{chung2018unsupervised}. However, in a number of sequence modeling tasks, the input and output sequences are of different lengths (e.g., speech recognition, neural machine translation, etc.), and the use of conventional RNNs in these tasks may be suboptimal.

To address these tasks, three main approaches including the use of attention-based encoder-decoders (AEDs) \cite{bahdanau2015neural, kim2017joint, li2022recent}, connectionist temporal classification (CTC) loss \cite{graves2006connectionist}, and RNN-transducers (RNN-Ts) \cite{2012Sequence} have been largely developed. However, not all of them are equally suited to all sequence modeling tasks. For example, the order of the input and output sequences will always be preserved for tasks such as speech recognition and handwriting recognition \cite{lu17b_interspeech}, but AEDs cannot guarantee this order preservation constraint will always be met (unlike CTC). Similarly, both CTC and RNN-Ts share a common limitation of having discontinuous and sparse latent spaces.
This becomes a problem when the test data is mapped to unexplored area of the latent space. Although variational modeling has been successfully applied in various domains (e.g., data generation) to address the discontinuity and sparsity issues \cite{kingma2013auto, sohn2015learning}, so far, it has not been incorporated with CTC. 
The challenge mainly arises from approximating the joint distribution of the latent variable sequence across all time steps, rendering the standard form of the variational lower bound and the model log-likelihood intractable.

In this paper, we present a novel approach to order-preserving sequence modeling, where CTC is incorporated with variational modeling. The proposed approach aims to leverage both the strength of CTC, which guarantees to preserve order, and that of variational modeling, which does not produce discontinuous and sparse latent spaces. To tackle the intractability of the variational lower bound, we show that one of two pragmatic assumptions can be made. Specifically, the first assumption is that the variational latent variables at each time step are conditionally independent, while the second assumes the temporal dependency amongst the latent variables is Markovian. Through mathematical derivations, both assumptions lead to tractable loss functions, allowing direct optimization of the variational lower bound of the model log-likelihood. Errors caused by the discontinuity and sparsity of the latent space can thus be mitigated.


\section{Order-preserving Sequence Modeling}
A general sequence modeling problem can be formally described as:
\begin{problem}\label{problem_sequence_modeling}
{\it The input sequence and the corresponding target sequence are represented as $\mathbf{X} = \left [ \mathbf{x}_1, \ldots, \mathbf{x}_{T_{\text{in}}} \right ]$ and $y = [y_1, \ldots,$ $y_{T_{\text{out}}} ]$, respectively. For all $t$, $\mathbf{x}_t \in \mathbb{R}^{D_\text{in}}$, and $y_t$ comes from the target vocabulary $\mathcal{V}$. The task is to find a model $\mathcal{M}$ that maps the input sequence $\mathbf{X}$ to the target sequence $y$.
}
\end{problem}

All the three aforementioned approaches to tackling Problem~\ref{problem_sequence_modeling} can be understood in the context of systems abstracted as:
\begin{equation}\label{eq.deterministic}
\mathbf{Z} = f(\mathbf{X}), \quad y = g(\mathbf{X}, \mathbf{Z}),
\end{equation}
where the input sequence, $\mathbf{X} \in \mathbb{R}^{D_{\text{in}} \times T_{\text{in}}}$, is first mapped into a latent variable sequence, $\mathbf{Z} = [\mathbf{z}_1, \ldots, \mathbf{z}_{T_{\text{in}}}]$, by $f$, and the target sequence $y$ obtained using another mapping function $g$. 
AEDs output a token, $y_{t_{\text{out}}}$, conditioned on the entire sequence $\mathbf{Z}$ and the history of the target token $y_{< {t_{\text{out}}}} := \left [ y_1, \ldots, y_{{t_{\text{out}}} - 1} \right ]$, i.e., it predicts $p \left ( y_{t_{\text{out}}} | \mathbf{h}, y_{< {t_{\text{out}}}} \right )$ at each time step, where $\mathbf{h}$ is a vector further abstracted from $\mathbf{Z}$ (e.g., pooling average of the hidden states of an RNN-based encoder) \cite{bahdanau2015neural}. However, for tasks like speech recognition or handwriting recognition, an additional constraint that $\mathcal{M}$ is {\it order-preserving} should also be imposed. That is to say, a piece of semantic should appear at the identical relative location within the input and target sequences \cite{lu17b_interspeech}, i.e., 
\begin{definition}\label{def.order_preserving}
{\it Given $y_{t_{\text{out}}^{(1)}} = g( \mathbf{X}_{t_{\text{in}}^{(1)}}, \mathbf{Z}_{t_{\text{in}}^{(1)}} )$ and $y_{t_{\text{out}}^{(2)}} = g( \mathbf{X}_{t_{\text{in}}^{(2)}},$ $\mathbf{Z}_{t_{\text{in}}^{(2)}} )$, for all $t_{\text{in}}^{(1)} \leq t_{\text{in}}^{(2)}$, if $t_{\text{out}}^{(1)} \leq t_{\text{out}}^{(2)}$ always holds, then $\mathcal{M}$ is order-preserving. $t_{\text{in}}^{(1)}$ (or $t_{\text{in}}^{(2)}$) could denote the last time step of the input sequence if multiple input time steps map to a single output time step $t_{\text{out}}^{(1)}$ (or $t_{\text{out}}^{(2)}$); and vice versa.}
\end{definition}

\begin{figure}[]
\centering
\includegraphics[width=0.4\textwidth]{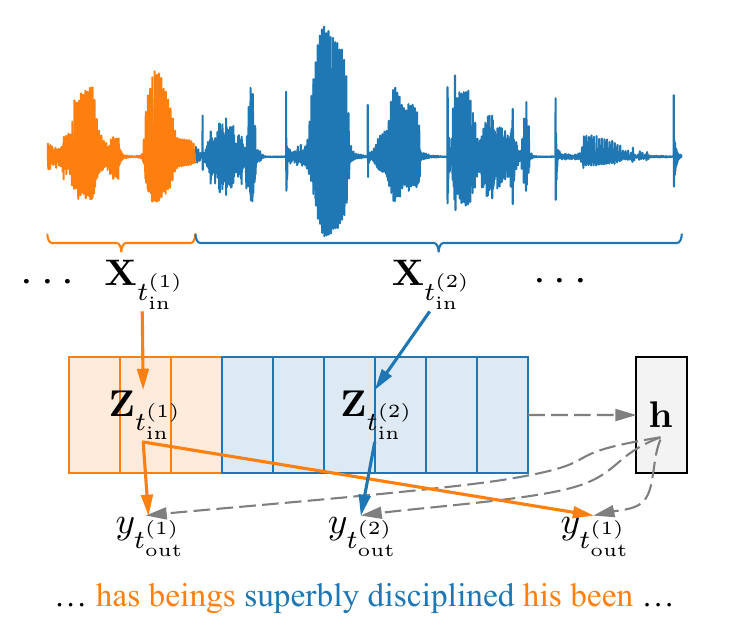}
\caption{A speech recognition example illustrating that AEDs are not guaranteed to preserve order based on an AED system \cite{chan2015listen}. The ground truth is $y_{t_{\text{out}}^{(1)}} =$ ``has been'' and $y_{t_{\text{out}}^{(2)}} =$ ``superbly disciplined''. However, due to the loss of ordering information, $y_{t_{\text{out}}^{(2)}}$ is emitted before $y_{t_{\text{out}}^{(1)}}$ in the predicted sequence by the AED.}\label{fig.aed_vs_ctc} 
\end{figure}

While AEDs can achieve order-aware modeling to a certain extent upon learning sequence alignments via their attention mechanisms, the lack of explicit constraints (i.e., Definition~\ref{def.order_preserving}) on the modeled alignment between $\mathbf{X}$ and $y$ can lead to the violation of $t_{\text{in}}^{(1)} \leq t_{\text{in}}^{(2)} \; \Rightarrow \; t_{\text{out}}^{(1)} \leq t_{\text{out}}^{(2)}$. To be specific, even though $\mathbf{h}$ is abstracted from the entire input sequence, the vital ordering information has been lost. As illustrated by a speech recognition example in Fig.~\ref{fig.aed_vs_ctc}, when decoding the target sequence with $\mathbf{h}$, the decoder could emit $y_{t_{\text{out}}^{(2)}}$ before $y_{t_{\text{out}}^{(1)}}$, violating Definition~\ref{def.order_preserving}. Furthermore, the performance of AEDs over long sequences is also easily impacted by noise \cite{chorowski2015attention}, and they can suffer large time delays during decoding \cite{chan2016listen}.

Unlike AEDs which decode the target sequence given the entire encoded sequence, the CTC approach and RNN-Ts predict the target sequence token by token as $p \left ( y_{t_{\text{out}}} | \mathbf{Z}_{\leq t_{\text{in}}} \right )$ and $p \left ( y_{t_{\text{out}}} | \mathbf{Z}_{\leq t_{\text{in}}}, y_{< t_{\text{out}}} \right )$, respectively, where $\mathbf{Z}_{\leq t_{\text{in}}} := \left [\mathbf{z}_1, \ldots, \mathbf{z}_{t_{\text{in}}} \right ]$, leading to an intrinsic order-preserving constraint \cite{graves2006connectionist, 2012Sequence, lu17b_interspeech}. Even though RNN-Ts take into account the dependencies amongst the target sequence (i.e., information from $y_{< t_{\text{out}}}$ is used when predicting $y_{t_{\text{out}}}$), it is difficult for this model to converge, as the prediction network usually requires pre-training \cite{hu2020exploring}. The CTC is easier to implement as it does not require a recurrent structure nor pre-training \cite{li2022recent}. The CTC loss optimizes the log-likelihood of the target sequence $\log p(y | \mathbf{X})$ by marginalizing over the set of all possible alignments, i.e., 
\begin{equation}\label{eq.ctc_loss}
\tilde{\mathcal{L}}_{\text{CTC}} = \sum_{A \in F^{-1}(y)} \prod_{t = 1}^{T_{\text{in}}} p(a_t | \mathbf{X}), \quad a_t \in \mathcal{V} \cup \{ - \}, \forall t,
\end{equation}
where $A = [a_1, \ldots, a_{T_{\text{in}}}]$ denotes an alignment between $\mathbf{X}$ and $y$, and is usually referred to as a {\it path} \cite{graves2006connectionist}, with $a_t$ representing the output at time step $t$. $\mathcal{V}$ is the target vocabulary. The many-to-one mapping function $F$ maps the paths $A$ to the target sequence $y$ by first merging the consecutive duplicated labels into one and then discarding the blanks ``$-$''.


\begin{figure}[]
\centering
\includegraphics[width=0.45\textwidth]{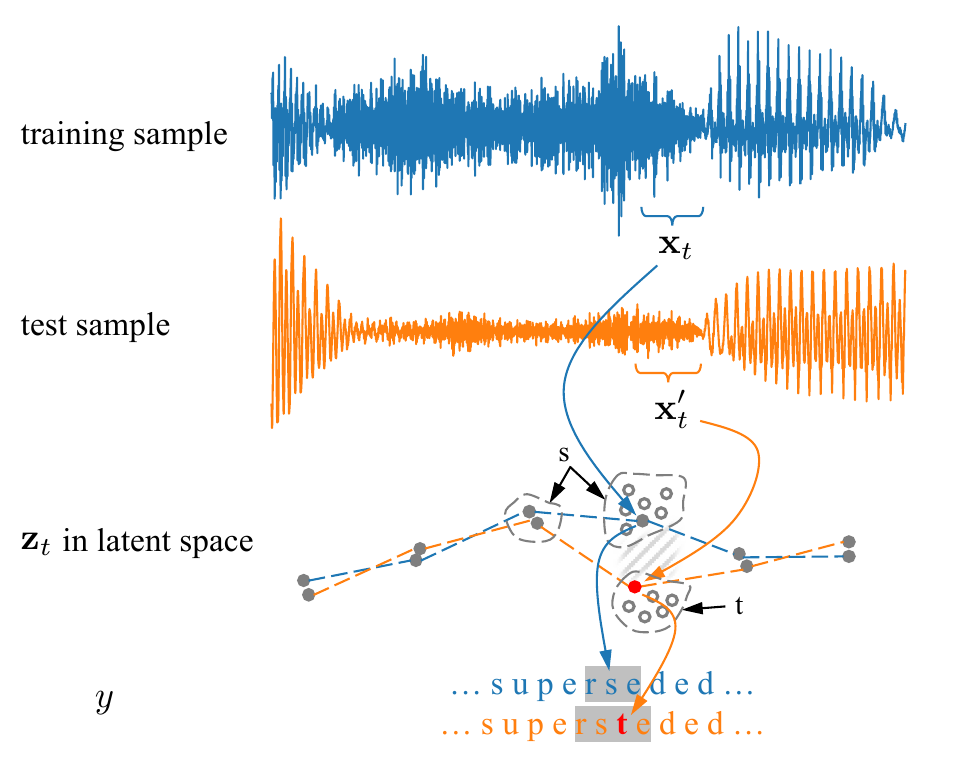}
\caption{A speech recognition example demonstrating an error caused by discontinuity and sparsity of latent space based on a deterministic sequence model \cite{baevski2020wav2vec}.}\label{fig.ctc_vs_vctc}
\end{figure}

Finally, CTC, as well as the other two approaches (AEDs and RNN-Ts), are only applied to {\it deterministic} sequence models given by \eqref{eq.deterministic}, where the latent space is discontinuous and sparse \cite{chung2015recurrent, kingma2013auto}, resulting in a certain type of inference error. 
This is illustrated by another speech recognition example in Fig.~\ref{fig.ctc_vs_vctc}. Let $\mathbf{x}_t$ denote the $t$-th feature vector from the input sequence $\mathbf{X}$ of a training sample, which corresponds to the token ``s'' in the target sequence ``... super\underline{s}eded ...''. 
During training, $\mathbf{x}_t$ is mapped to a single point in the latent sub-space, and feature vectors from the same class are mapped into one cluster, such as the cluster corresponding to ``s''. Given the discontinuous and sparse nature of the latent space, there are vast unexplored regions in the latent space that lack data coverage, as represented by the shaded grey area. This limitation of the model can lead to incorrect predictions when faced with unseen data, indicating insufficient robustness to data variability. 
For an $\mathbf{x}'_t$ from a test sequence $\mathbf{X}' \not\in \mathcal{D}$ close to $\mathbf{X}$, it may be mapped to the unexplored area in the latent sub-space and recognized as a point in a wrong cluster. For instance, ``s'' is mistakenly recognized as ``t'' in Fig.~\ref{fig.ctc_vs_vctc}. 


\section{Variational Sequence Modeling Preliminaries}\label{seq.preliminary}
Variational models deal with scenarios where the test data is not mapped to explored area of the latent space (as illustrated by Fig.~\ref{fig.ctc_vs_vctc}) by mapping the input data to collections of distributions in the latent space (instead of points). This enables a more comprehensive and compact exploration of the latent space and a richer representation of each cluster. By capturing the uncertainty in the latent space, variational modeling improve the model generalization, making the model more adaptable to diverse and unseen data. It can be represented as
\begin{equation}\label{eq.variational_model}
\left \{
\begin{aligned}
\mathbf{Z} & \sim p(\mathbf{Z} | \mathbf{X}; \theta), \text{ where } \theta = f_v(\mathbf{X}), \\
y & = g_v(\mathbf{X}, \mathbf{Z}). \\
\end{aligned}
\right .
\end{equation}
Different from \eqref{eq.deterministic}, the input $\mathbf{X}$ of \eqref{eq.variational_model} is first transformed into a continuous distribution $p(\mathbf{Z} | \mathbf{X}; \theta)$ via $f_v$. A latent variable sequence $\mathbf{Z} \in \mathbb{R}^{D_{\mathbf{z}} \times T_{\text{in}}}$ is then drawn from this distribution, and subsequently mapped to $y$ via $g_v$, where $D_{\mathbf{z}}$ is the dimension of $\mathbf{z}_t$. 
Since $\mathbf{X}$ can be mapped to any point within an area governed by $p(\mathbf{Z} | \mathbf{X}; \theta)$ instead of one particular point, the latent space is largely explored during training and as a result much more compact.
The variational extensions to AEDs, collectively referred to as variational encoder-decoders (VEDs), show better performance when compared to vanilla AEDs \cite{bahuleyan2018variational, zhao2021transformer, li2017deep}. They generally include the variational latent variables at the last hidden state of the encoder \cite{bahuleyan2018variational}, the attention mechanism \cite{bahuleyan2018variational}, the context vector \cite{zhao2021transformer}, and the estimate of the output \cite{li2017deep} in existing studies. However, all the shortcomings of AEDs (not guaranteed to order-preserving, sensitive to noise, longer time delay, etc.) that we mentioned above still exist in VEDs.

Due to the order-preserving character of CTC and the generalization offered by variational modeling, there is a compelling rationale for a variational version of CTC. However, 
it turns out that in general such an objective function for model development is intractable.
To be specific, in order to maximize the log-likelihood of a variational sequence model, the variational lower bound $\mathcal{L}$
\begin{equation}\label{eq.likelihood}
\begin{aligned}
& \log p( y | \mathbf{X} ) \\
= & \mathbb{E}_{q( \mathbf{Z} | \mathbf{X})} \left [ \log \frac{ p( y | \mathbf{X}, \mathbf{Z} ) p( \mathbf{Z} | \mathbf{X} ) q( \mathbf{Z} | \mathbf{X} ) }{ p( \mathbf{Z} | \mathbf{X}, y ) q( \mathbf{Z} | \mathbf{X} ) } \right ] \\
= & \mathbb{E}_{q( \mathbf{Z} | \mathbf{X})} \left [ \log p( y | \mathbf{X}, \mathbf{Z} ) \right ] - \text{KL} ( q( \mathbf{Z} | \mathbf{X} ) \| p( \mathbf{Z} | \mathbf{X} ) ) \\
& \quad + \text{KL} ( q( \mathbf{Z} | \mathbf{X} ) \| p( \mathbf{Z} | \mathbf{X}, y ) ) \\
\geq & \underbrace{ \mathbb{E}_{q(\mathbf{Z} | \mathbf{X})}[ \log p(y | \mathbf{X}, \mathbf{Z}) ] }_{\text{prediction term}} - \underbrace{ \text{KL} ( q(\mathbf{Z} | \mathbf{X}) \| p(\mathbf{Z} | \mathbf{X}) ) }_{\text{regularization term}} = \mathcal{L}
\end{aligned}
\end{equation}
is typically optimized \cite{sohn2015learning}. As shown above, $\mathcal{L}$ consists of a prediction term and a regularization term, with $\text{KL}(\cdot \| \cdot)$ denoting the Kullback-Leibler (KL) divergence, and $\text{KL} ( q( \mathbf{Z} | \mathbf{X} ) \|$ $p( \mathbf{Z} | \mathbf{X}, y ) ) \geq 0$. Here, $q( \mathbf{Z} | \mathbf{X})$ is a distribution that approximates the posterior $p( \mathbf{Z} | \mathbf{X}, y )$. The variational lower bound $\mathcal{L}$ is generally a tight lower bound.
As mentioned in Section~\ref{sec.introduction}, approximation of the joint distribution of $\mathbf{Z} = [\mathbf{z}_1, \ldots, \mathbf{z}_{T_{\text{in}}}]$ over all time steps, and subsequently $\mathcal{L}$, are both intractable. Nonetheless, with certain assumption, $\mathcal{L}$ can be simplified and made tractable.  


\section{Proposed Variational Connectionist Temporal Classification}\label{sec.variational_ctc}
In this section, we transform the standard form of variational lower bound $\mathcal{L}$ in \eqref{eq.likelihood} into tractable forms based on two commonly adopted assumptions in sequence modeling, i.e., conditional independence and Markovian dependence. The two resultant loss functions are respectively developed in the following two subsections.


\subsection{Conditional Independence}\label{sec.conditional_independence}
We first consider the case where the latent variables $\mathbf{z}_t, \forall t$ corresponding to each time step are generated independently, conditioned on the input sequence $\mathbf{X}$. 
\begin{assumption}\label{assumption_1}
{\it Given the input sequence $\mathbf{X} \in \mathbb{R}^{D_{\text{in}} \times T_{\text{in}}}$, the latent variables $\mathbf{z}_t \in \mathbb{R}^{D_{\mathbf{z}}}, t = 1, \ldots, T_{\text{in}}$ drawn from $p(\mathbf{Z} | \mathbf{X})$ are conditionally independent, i.e.,
\begin{equation}\label{eq.conditional_independence}
p(\mathbf{Z} | \mathbf{X}) = \prod_{t = 1}^{T_{\text{in}}} p(\mathbf{z}_t | \mathbf{X}).
\end{equation}
}
\end{assumption}

Based on Assumption~\ref{assumption_1}, the regularization term in \eqref{eq.likelihood} can be first decomposed as below:
\begin{align}
\nonumber & \text{KL} ( q(\mathbf{Z} | \mathbf{X}) \| p(\mathbf{Z} | \mathbf{X}) ) \\
\nonumber = & \int_{ \{ \mathbf{z}_t \}_{t = 1}^{T_{\text{in}}} } \prod_{t = 1}^{T_{\text{in}}} q(\mathbf{z}_t | \mathbf{X}) \cdot \log \frac{ \prod_{t = 1}^{T_{\text{in}}} q(\mathbf{z}_t | \mathbf{X}) }{ \prod_{t = 1}^{T_{\text{in}}} p(\mathbf{z}_t | \mathbf{X}) } d \mathbf{z}_{1: T_{\text{in}}} \\
\nonumber = & \int_{ \{ \mathbf{z}_t \}_{t = 1}^{T_{\text{in}}} } \prod_{t = 1}^{T_{\text{in}}} q(\mathbf{z}_t | \mathbf{X}) \cdot \left [ \sum_{t = 1}^{T_{\text{in}}} \log \frac{ q(\mathbf{z}_t | \mathbf{X}) }{ p(\mathbf{z}_t | \mathbf{X}) } \right ] d \mathbf{z}_{1: T_{\text{in}}} \\
\nonumber = & \sum_{t = 1}^{T_{\text{in}}} \int_{ \{ \mathbf{z}_t \}_{t = 1}^{T_{\text{in}}} } \prod_{\tau = 1}^{T_{\text{in}}} q(\mathbf{z}_{\tau} | \mathbf{X}) \cdot \log \frac{ q(\mathbf{z}_{t} | \mathbf{X}) }{ p(\mathbf{z}_{t} | \mathbf{X}) } d \mathbf{z}_{1: T_{\text{in}}} \\
\nonumber = & \sum_{t = 1}^{T_{\text{in}}} \int_{\mathbf{z}_t} q(\mathbf{z}_t | \mathbf{X}) \log \frac{ q(\mathbf{z}_t | \mathbf{X}) }{ p(\mathbf{z}_t | \mathbf{X}) } d \mathbf{z}_t \prod_{\substack{\tau = 1 \\ \tau \neq t}}^{T_{\text{in}}} \int_{\mathbf{z}_{\tau}} q(\mathbf{z}_{\tau} | \mathbf{X}) d \mathbf{z}_{\tau} \\
\nonumber = & \sum_{t = 1}^{T_{\text{in}}} \int_{\mathbf{z}_t} q(\mathbf{z}_t | \mathbf{X}) \log \frac{ q(\mathbf{z}_t | \mathbf{X}) }{ p(\mathbf{z}_t | \mathbf{X}) } d \mathbf{z}_t \\
\label{eq.kl_derivation_conditional}= & \sum_{t = 1}^{T_{\text{in}}} \text{KL} ( q(\mathbf{z}_t | \mathbf{X}) \| p(\mathbf{z}_t | \mathbf{X}) ),
\end{align}
where $\int_{\mathbf{z}_{\tau}} q(\mathbf{z}_{\tau} | \mathbf{X}) d \mathbf{z}_{\tau} = 1$ for all $\tau = 1, \ldots, T_{\text{in}}, \tau \neq t$, and thus their product is equal to $1$.
Note that in \eqref{eq.kl_derivation_conditional}, we let the approximate distribution $q(\mathbf{Z} | \mathbf{X})$ be factorizable, i.e., $q(\mathbf{Z} | \mathbf{X}) = \prod_{t = 1}^{T_{\text{in}}} q(\mathbf{z}_t | \mathbf{X})$. The regularization term thus can be decomposed into a series of KL divergences corresponding to each time step $t$. 
In other words, it is no longer necessary to model the joint distribution $p(\mathbf{Z} | \mathbf{X})$ or $q(\mathbf{Z} | \mathbf{X})$ of the whole sequence, but simply define the distributions $p(\mathbf{z}_t | \mathbf{X})$ and $q(\mathbf{z}_t | \mathbf{X})$ at each time step $t$, which can be approximated by appropriate prior distributions (e.g., Gaussian or Bernoulli).

Next, for the prediction term of \eqref{eq.likelihood}, we can use Monte Carlo sampling to estimate the expectation of $\log p(y | \mathbf{X}, \mathbf{Z})$ w.r.t. $q(\mathbf{Z} | \mathbf{X})$ as:
\begin{equation}\label{eq.monte_carlo_estimate}
\mathbb{E}_{q(\mathbf{Z} | \mathbf{X})} \left [ \log p(y | \mathbf{X}, \mathbf{Z}) \right ] \simeq \frac{1}{L} \sum_{l = 1}^L \log p \left ( y \left | \mathbf{X}, \mathbf{Z}^{(l)} \right ) \right . , 
\end{equation}
where $\mathbf{Z}^{(l)}$ represents a sequence of length $T_{\text{in}}$ sampled from $q(\mathbf{Z} | \mathbf{X})$, and $L$ is the number of sampled sequences.
Although a larger $L$ (i.e., more samples) generally leads to a better approximation in \eqref{eq.monte_carlo_estimate}, it has been suggested that $L$ can be set to 1 as long as the size of the mini-batch is large enough \cite{kingma2013auto}. This not only reduces the computation complexity without sacrificing accuracy, but also simplifies \eqref{eq.monte_carlo_estimate} to: 
\begin{equation}\label{eq.reconstruction}
\mathbb{E}_{q(\mathbf{Z} | \mathbf{X})} \left [ \log p(y | \mathbf{X}, \mathbf{Z}) \right ] \simeq \log p(y | \mathbf{X}, \mathbf{Z}),
\end{equation}
which can be optimized by the standard CTC loss $\tilde{\mathcal{L}}_{\text{CTC}}$ in \eqref{eq.ctc_loss}. With Assumption~\ref{assumption_1}, $\forall t$, $a_t$ is only dependent on $\mathbf{X}$ and $\mathbf{z}_t$, i.e., $p(A | \mathbf{X}, \mathbf{Z}) = \prod_{t = 1}^{T_{\text{in}}} p(a_t | \mathbf{X}, \mathbf{z}_t)$. Recall that this is just the conditional independence assumption \cite{graves2006connectionist} required by the standard CTC approach \eqref{eq.ctc_loss}. Therefore, order preservation can be guaranteed, and combining \eqref{eq.likelihood}, \eqref{eq.kl_derivation_conditional} and \eqref{eq.reconstruction}, we propose:
\begin{proposition}\label{lemma_1}
{\it Given an input sequence $\mathbf{X}$ and the target sequence $y$ with different lengths, 
a variational model which introduces latent variable sequence $\mathbf{Z}$ can be optimized with the loss function 
}
\begin{equation}\label{eq.loss_conditional_independence}
\tilde{\mathcal{L}}_c = \sum_{A \in {F^{-1}(y)}} \prod_{t = 1}^{T_{\text{in}}} p(a_t | \mathbf{X}, \mathbf{Z}) - \sum_{t = 1}^{T_{\text{in}}} \text{KL} ( q(\mathbf{z}_t | \mathbf{X}) \| p(\mathbf{z}_t | \mathbf{X}) ).
\end{equation}
\end{proposition}


Although the prior and posterior distributions of $\mathbf{z}_t$ can take any form, they are often assumed to be Gaussian distributions \cite{kingma2013auto}. This assumption also leads to closed-form expressions for the KL divergences in \eqref{eq.loss_conditional_independence}. Consequently, we also provide a closed-from expression under this Gaussianity assumption here. Upon letting both $q(\mathbf{z}_t | \mathbf{X})$ and $p(\mathbf{z}_t | \mathbf{X})$ be defined as multivariate Gaussians with diagonal covariances, i.e., $\mathcal{N} \left ( \boldsymbol{\mu}_t, \boldsymbol{\sigma}_t^2 \mathbf{I} \right )$ and $\mathcal{N} \left ( \check{\boldsymbol{\mu}}_t, \check{\boldsymbol{\sigma}}_t^{2} \mathbf{I} \right )$, respectively \cite{kingma2013auto}, the KL divergences take on the following closed-form expression
:
\begin{equation}\label{eq.kl_closed_form}
\begin{aligned}
& \text{KL} ( q(\mathbf{z}_t | \mathbf{X}) \| p(\mathbf{z}_t | \mathbf{X}) ) \\
= & - \frac{1}{2} \sum_{d = 1}^{D_{\mathbf{z}}} \left [ \log \frac{\sigma_{t, d}^2}{\check{\sigma}_{t, d}^{2}} - \frac{\sigma_{t, d}^2}{\check{\sigma}_{t, d}^{2}} - \frac{\left ( \mu_{t, d} - \check{\mu}_{t, d} \right )^2}{\check{\sigma}_{t, d}^{2}} + 1 \right ], 
\end{aligned}
\end{equation}
where $\boldsymbol{\mu}_t, \boldsymbol{\sigma}_t, \check{\boldsymbol{\mu}}_t, \check{\boldsymbol{\sigma}}_t \in \mathbb{R}^{D_{\mathbf{z}}}$ are the parameters of the two distributions, and they can be either pre-defined or learned jointly using neural networks. With \eqref{eq.kl_closed_form} substituted into \eqref{eq.loss_conditional_independence}, the variational lower bound in \eqref{eq.likelihood} can thus be effectively improved by maximizing this tractable loss function $\tilde{\mathcal{L}}_c$. 

\begin{figure}[]
\centering
\includegraphics[width=0.48\textwidth]{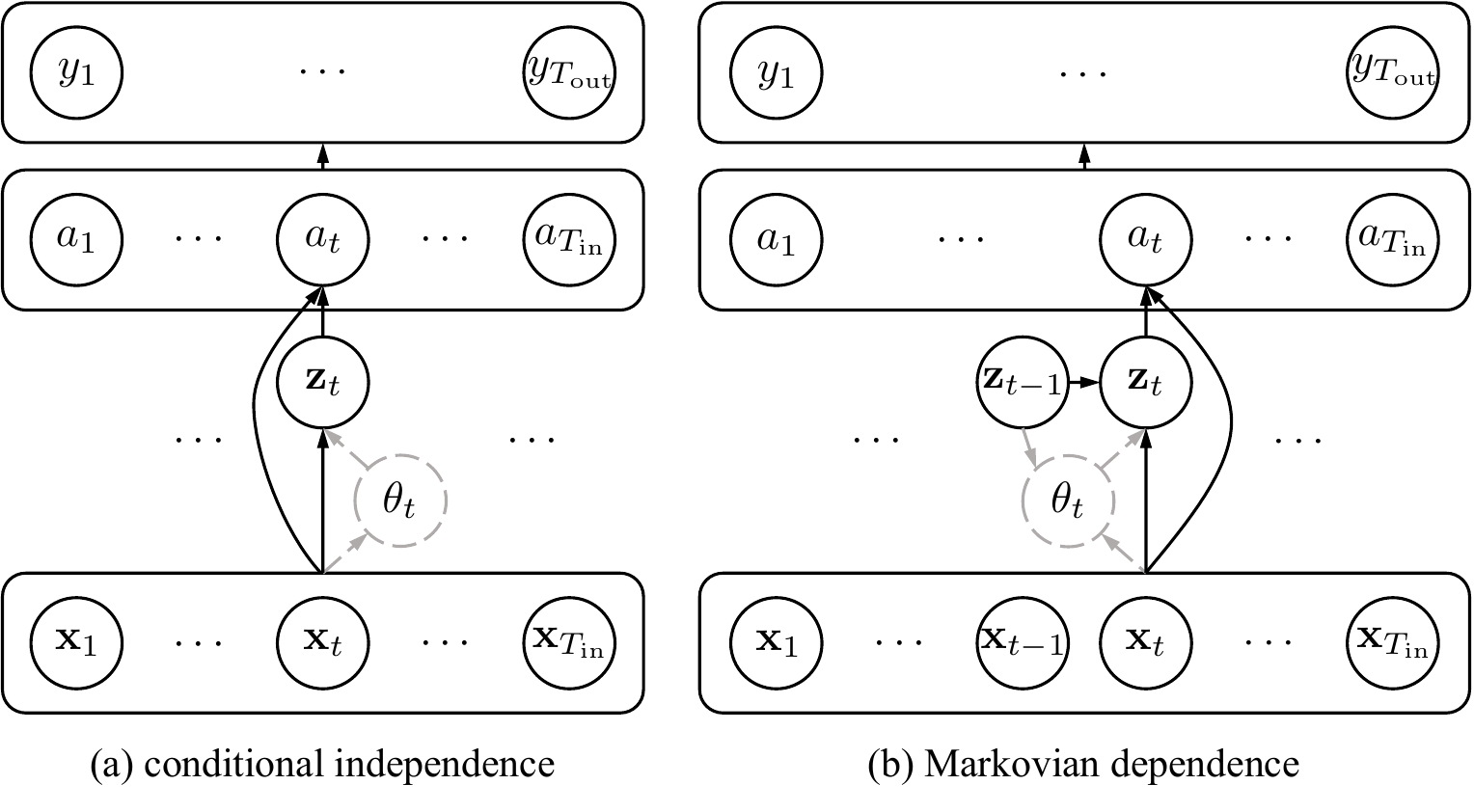}
\caption{Illustrations of dependencies among variables under (a)~conditional independence and (b)~Markovian dependence assumptions. Specifically, $\theta_t = (\boldsymbol{\mu}_t, \boldsymbol{\sigma}_t)$ when $q(\mathbf{z}_t | \mathbf{X}) = \mathcal{N} \left ( \boldsymbol{\mu}_t, \boldsymbol{\sigma}_t^2 \mathbf{I} \right )$. 
}\label{fig.model_structure}
\end{figure}

It is worth mentioning that there exists a special case where Assumption~\ref{assumption_1} is already made in the model. 
To be more specific, for variational models developed in many instances (e.g., \cite{huang2020deep}), every $\mathbf{z}_t$ is independently learned only from $\mathbf{X}$ or $\mathbf{x}_t$, as illustrated in Fig.~\ref{fig.model_structure}~(a). This design is primarily intended to facilitate parallel computation during model training \cite{wang2020survey}. As a result, Assumption~\ref{assumption_1} is already met in such cases. The loss function given by Proposition~\ref{lemma_1} could precisely optimize the variational lower bound for the log-likelihood in this case.

\subsection{Markovian Dependence}\label{sec.markov}
In this subsection, we consider a more complex case where the temporal dependency over time sequence $\mathbf{Z}$ is incorporated. Specifically, when there is non-negligible dependency among the latent variables $\mathbf{z}_t$, it can be assumed to be Markovian \cite{gagniuc2017markov}. This model is depicted in Fig.~\ref{fig.model_structure}~(b).
\begin{assumption}\label{assumption_2}
{\it The latent variables $\mathbf{z}_t, t = 1, \ldots, T_{\text{in}}$ in model $\mathcal{M}$ satisfy the Markov assumption, i.e.,
\begin{equation}\label{eq.markov}
p(\mathbf{X}, \mathbf{Z}) = p(\mathbf{X}) \prod_{t = 1}^{T_{\text{in}}} p(\mathbf{z}_t | \mathbf{z}_{t - 1}, \mathbf{X}),
\end{equation}
where, for $t = 1$, $p(\mathbf{z}_1 | \mathbf{z}_{0}, \mathbf{X}) = p(\mathbf{z}_1 | \mathbf{X})$.
}
\end{assumption}

The corresponding loss function can then be derived, leading to the following proposition:
\begin{proposition}\label{lemma_2}
{\it Taking into account the temporal dependency among the latent variables $\mathbf{z}_t$, we can optimize the model $\mathcal{M}$ via the loss function:
}
\begin{equation}\label{eq.lb_markov}
\begin{aligned}
\tilde{\mathcal{L}}_m = & \sum_{A \in F^{-1}(y)} \prod_{t = 1}^{T_{\text{in}}} p(a_t | \mathbf{X}, \mathbf{Z}) \\
& \quad - \sum_{t = 1}^{T_{\text{in}}} \mathbb{E}_{ q(\mathbf{z}_{t - 1} | \mathbf{X}) } \left [ \text{KL} ( q(\mathbf{z}_t | \mathbf{X}) \| p(\mathbf{z}_t | \mathbf{z}_{t - 1}, \mathbf{X}) ) \right ].
\end{aligned}
\end{equation}
\end{proposition}

\begin{proof}
Similar to \eqref{eq.kl_derivation_conditional}, we still let $q(\mathbf{Z} | \mathbf{X})$ be factorized as $q(\mathbf{Z} | \mathbf{X}) = \prod_{t = 1}^{T_{\text{in}}} q(\mathbf{z}_t | \mathbf{X})$. Then, with Assumption~\ref{assumption_2}, we have $(\mathbf{Z} | \mathbf{X}) = p(\mathbf{X}, \mathbf{Z}) / p(\mathbf{X}) = \prod_{t = 1}^{T_{\text{in}}} p(\mathbf{z}_t | \mathbf{z}_{t - 1}, \mathbf{X})$. Following \eqref{eq.kl_derivation_conditional}, we derive the regularization term in \eqref{eq.likelihood} as
\begin{align}
\nonumber & \text{KL} ( q(\mathbf{Z} | \mathbf{X}) \| p(\mathbf{Z} | \mathbf{X}) ) \\
\nonumber = & \int_{ \{ \mathbf{z}_t \}_{t = 1}^{T_{\text{in}}} } \prod_{t = 1}^{T_{\text{in}}} q(\mathbf{z}_t | \mathbf{X}) \cdot \left [ \sum_{t = 1}^{T_{\text{in}}} \log \frac{ q(\mathbf{z}_t | \mathbf{X}) }{ p(\mathbf{z}_t | \mathbf{z}_{t - 1}, \mathbf{X}) } \right ] d \mathbf{z}_{1: T_{\text{in}}} \\
\nonumber = & \sum_{t = 1}^{T_{\text{in}}} \int_{ \{ \mathbf{z}_t \}_{t = 1}^{T_{\text{in}}} } \prod_{\tau = 1}^{T_{\text{in}}} q(\mathbf{z}_{\tau} | \mathbf{X}) \cdot \log \frac{ q(\mathbf{z}_{t} | \mathbf{X}) }{ p(\mathbf{z}_{t} | \mathbf{z}_{t - 1}, \mathbf{X}) } d \mathbf{z}_{1: T_{\text{in}}} \\
\nonumber = & \sum_{t = 1}^{T_{\text{in}}} \int_{\mathbf{z}_{t - 1}, \mathbf{z}_t} q(\mathbf{z}_{t - 1} | \mathbf{X}) q(\mathbf{z}_t | \mathbf{X}) \cdot \\
\nonumber & \quad \quad \log \frac{ q(\mathbf{z}_t | \mathbf{X}) }{ p(\mathbf{z}_t | \mathbf{z}_{t - 1}, \mathbf{X}) } d \mathbf{z}_{t - 1} d \mathbf{z}_t \prod_{\substack{\tau = 1 \\ \tau \neq t - 1 \\ \tau \neq t}}^{T_{\text{in}}} \int_{\mathbf{z}_{\tau}} q(\mathbf{z}_{\tau} | \mathbf{X}) d \mathbf{z}_{\tau} \\
\nonumber = & \sum_{t = 1}^{T_{\text{in}}} \int_{\mathbf{z}_{t - 1}, \mathbf{z}_t} q(\mathbf{z}_{t - 1} | \mathbf{X}) q(\mathbf{z}_t | \mathbf{X}) \log \frac{ q(\mathbf{z}_t | \mathbf{X}) }{ p(\mathbf{z}_t | \mathbf{z}_{t - 1}, \mathbf{X}) } d \mathbf{z}_{t - 1} d \mathbf{z}_t \\
\label{eq.kl_derivation_markov} = & \sum_{t = 1}^{T_{\text{in}}} \mathbb{E}_{ q(\mathbf{z}_{t - 1} | \mathbf{X}) } \left [ \text{KL} ( q(\mathbf{z}_t | \mathbf{X}) \| p(\mathbf{z}_t | \mathbf{z}_{t - 1}, \mathbf{X}) ) \right ],
\end{align}
where $\int_{\mathbf{z}_{\tau}} q(\mathbf{z}_{\tau} | \mathbf{X}) d \mathbf{z}_{\tau} = 1$ for all $\tau = 1, \ldots, T_{\text{in}}, \tau \neq t - 1, \tau \neq t$.
Together with \eqref{eq.likelihood}, \eqref{eq.reconstruction} and \eqref{eq.kl_derivation_markov}, we eventually obtain the loss function with form \eqref{eq.lb_markov}.
\end{proof}

Again, the expectations w.r.t. the KL divergences in \eqref{eq.lb_markov} can be estimated via Monte Carlo sampling as $\sum_{l = 1}^L \text{KL} ( q(\mathbf{z}_t^{(l)} | \mathbf{X}) \|$ $ p(\mathbf{z}_t^{(l)} | \mathbf{z}_{t - 1}^{(l)}, \mathbf{X}))$, and under the Gaussian assumption, the KL divergences take the same form as \eqref{eq.kl_closed_form}.

\section{Conclusion}\label{sec.conclusion}
In this paper, we present theoretical derivations of two tractable loss functions under two reasonable assumptions 
for the variational latent variables. 
The proposed losses allow direct optimization of the variational lower bound for the model log-likelihood. 
Models optimized with the proposed losses will achieve continuous and compact latent spaces, which will in turn reduce errors that are easily produced by deterministic counterparts optimized with the standard CTC loss. Besides the closed-form \eqref{eq.kl_closed_form} obtained under Gaussian assumption, alternative distribution assumptions (e.g., Bernoulli) that lead to closed-form loss functions are also worth exploring.





\bibliographystyle{IEEEbib}
\bibliography{references}

\end{document}